%% file: Paper-1908.tex
\documentclass[runningheads]{llncs}
\usepackage{float}
\usepackage[T1]{fontenc}
\usepackage{booktabs}
\usepackage{amsmath}
\usepackage{amssymb}
\usepackage{graphicx}
\usepackage{multicol}
\usepackage{multirow}
\usepackage{adjustbox}
\usepackage{hyperref}
\usepackage{multicol}
\usepackage{multirow}
\usepackage{makecell}
\usepackage{pifont}% http://ctan.org/pkg/pifont
\newcommand{\cmark}{\ding{51}}%
\newcommand{\xmark}{\ding{55}}%
\hypersetup{
    colorlinks=true,
    linkcolor=blue,
    filecolor=magenta,      
    urlcolor=cyan,
    pdftitle={3D Graph Diffusion},
    pdfpagemode=FullScreen,
    }
\usepackage{orcidlink}
\usepackage{url}

\begin{document}
\title{3D Vessel Graph Generation Using\\ Denoising Diffusion}

\author{
    Chinmay~Prabhakar\thanks{Contributed equally} \inst{1} \orcidlink{0000-0002-1780-8108}\and
    Suprosanna~Shit$^\star$ \inst{1} \orcidlink{0000-0003-4435-7207} \and
    Fabio~Musio \inst{1, 2} \and
    Kaiyuan~Yang \inst{1} \and
    Tamaz~Amiranashvili \inst{1, 3}\orcidlink{0000-0001-8914-3427} \and
    Johannes~C.~Paetzold \inst{4}\orcidlink{0000-0002-4844-6955}\and
    Hongwei~Bran~Li\inst{5} \orcidlink{0000-0002-5328-6407}\and
  Bjoern~Menze\inst{1} \orcidlink{0000-0003-4136-5690}}

% index{Prabhakar, Chinmay}
% index{Shit, Suprosanna}
% index{Musio, Fabio}
% index{Yang, Kaiyuan}
% index{Amiranashvili, Tamaz}
% index{Paetzold, Johannes C.} 
% index{Li, Hongwei Bran} 
% index{Menze, Bjoern}

\authorrunning{Prabhakar and Shit et al.}
\institute{Department of Quantitative Biomedicine, University of Zurich, Switzerland \and
 Center for Computational Health, ZHAW, Switzerland \and
 Department of Computer Science, Technical University of Munich, Germany \and
 Department of Computing, Imperial College London, UK \and
 Athinoula A. Martinos Center, Harvard Medical School, USA
\email{chinmay.prabhakar@uzh.ch}}

\maketitle              % typeset the header of the contribution

\begin{abstract}
Blood vessel networks, represented as 3D graphs, help predict disease biomarkers, simulate blood flow, and aid in synthetic image generation, relevant in both clinical and pre-clinical settings. However, generating realistic vessel graphs that correspond to an anatomy of interest is challenging. Previous methods aimed at generating vessel trees mostly in an autoregressive style and could not be applied to vessel graphs with cycles such as capillaries or specific anatomical structures such as the Circle of Willis. Addressing this gap, we introduce the first application of \textit{denoising diffusion models} in 3D vessel graph generation. Our contributions include a novel, two-stage generation method that sequentially denoises node coordinates and edges. We experiment with two real-world vessel datasets, consisting of microscopic capillaries and major cerebral vessels, and demonstrate the generalizability of our method for producing diverse, novel, and anatomically plausible vessel graphs.

% \keywords{Denoising diffusion \and Graph generation \and Vessels\.}
\end{abstract}
\input{01_intro}

\input{02_rel_lit}
\input{03_method}
\input{04_exp}

\section{Conclusion}

Generating spatial vascular graphs is a clinically relevant, albeit challenging task. In this work, we propose a two-stage approach to generate graphs that better capture the vascular intricacies. To the best of our knowledge, this is the first application of a diffusion model to generate vascular graphs. Since the diffusion process is at the heart of our method, we can easily extend our method to generate graphs conditioned on, for instance, disease labels. We defer such extensions to future work. Most importantly, diffusion models have revolutionized generative modeling, outperforming other competitive methods, and we hope our method paves the way for more widespread adoption of diffusion models for vascular graph generation.

\begin{credits}
\subsubsection{\ackname} This work has been supported by the Helmut Horten Foundation. S. S. is supported by the GRC Travel Grant from UZH. F.M. is funded by the DIZH grant. H. B. Li is supported by an SNF postdoctoral mobility grant.
% H.B.L. is supported by an Nvidia GPU research grant.

\subsubsection{\discintname}
The authors have no competing interests to declare that are relevant to the content of this article.
\end{credits}

\bibliographystyle{splncs04}
\bibliography{Paper-1908}
\clearpage
\appendix
\input{05_supp}
\end{document}

%% file: 01_intro.tex
\section{Introduction}
\label{sec:intro}

Studying blood vessel networks as a 3D spatial graph offers a compact representation of anatomical and physiological properties of the circulatory system and, hence, has gained increasing interest in clinical \cite{lyu2022reta} and pre-clinical image analysis \cite{paetzold2021whole}. Vessel graphs previously have been used to predict disease biomarkers \cite{paetzold2023geometric}, simulate blood flow \cite{reichold2009vascular}, and create synthetic image-label pairs for segmentation \cite{todorov2020machine,menten2022physiology}. However, annotating large-scale samples is expensive \cite{yang2023benchmarking} and tedious \cite{todorov2020machine}. To mitigate this issue, synthetic vessel graphs can augment real datasets for downstream tasks. While extracting vessel graphs from a segmented volume has been studied numerous times \cite{drees2021scalable,paetzold2021whole}, generating realistic vessel graphs remains relatively underexplored. Previous methods \cite{schneider2012tissue,menten2022physiology} primarily rely on rule-based vessel-tree generation based on a predefined oxygen concentration or nutrition distribution. However, such methods require precise domain modeling and often require tuning a substantial hyperparameter space. Therefore, a notable void exists, which is the generation of realistic 3D vessel graphs in an entirely data-driven way.

Recently, diffusion model-based graph generation has been successfully applied to various graph generation tasks ranging from molecule generation \cite{vignac2022digress}, protein design \cite{yi2024graph}, and material sciences \cite{xie2021crystal}. While these spatial graph generation tasks share a common goal, they differ significantly from each other in terms of the inductive bias describing the physical quantity of interest and how these quantities influence each other. For example, in the case of molecule generation, two neighboring atoms (nodes) being carbons provide a strong bias for the existence of a covalent bond. Such inductive bias in the form of node class, however, does not apply to vessel generation. Hence, the model has to learn entirely different co-dependencies and distributions; therefore, existing spatial graph generation methods are not readily applicable to vessel graph generation tasks. Furthermore, depending on the imaging resolutions, vessel graphs can have different topological and physiological properties. For example, capillaries are visible in microscopic images and contain a substantially higher number of cycles than large arteries, which are visible in magnetic resonance images of the brain. Hence, we aim to develop a flexible data-driven model capable of generating vessel graphs across different imaging resolutions.

In this work, we introduce the first denoising diffusion-based method for generating vessel graphs, addressing the challenge of simultaneously generating nodes and edges -- an issue we identify as ill-posed for vessel graphs. Our approach focuses on first generating node coordinates, followed by the denoising of edges. This strategy allows the network to accurately model node distribution before learning to connect these nodes into a coherent vascular graph topology. Our experiments demonstrate the capability of our method to produce valid 3D vessel graphs ranging from capillary to large vessel levels using real vessel data.

To summarize, our contributions are threefold. Firstly, we introduce a novel diffusion denoising-based vessel graph generation method, marking a first in this domain. Secondly, we address the complexity of generating vessel graphs by proposing a tailored approach, sequentially denoising nodes and then edges. Lastly, through experimentation with two real-world vessel datasets -- one representing capillaries and the other large vessels -- we validate our method's ability to generate diverse, unique, and valid vascular graphs.

%% file: 02_rel_lit.tex
\section{Related Literature}
\label{sec:rel_lit}

\paragraph{\textbf{Vessel Graph Generation:}}
Previous methods have investigated autoregressive vessel tree generation methods. Schneider et al.\cite{schneider2012tissue} used underlying oxygen concentration to generate an arterial tree. However, such simulations are costly, and recently, Rauch et al. \cite{rauch2021interactive} proposed a computationally efficient statistical algorithm. The resultant graphs are useful for generating synthetic image-segmentation pairs to train segmentation models \cite{kreitner2024synthetic,menten2022physiology}. Parallelly, there have been attempts to generate vessel mesh. Wolternick et al. \cite{wolterink2018blood} proposed a generative adversarial network to synthesize coronary arteries. Recently, Feldman et al. \cite{feldman2023vesselvae} used a variational-auto-encoder to generate intracranial vessel segments.
However, these network generation algorithms can not generate cycles in vessel graphs, which are crucial features for many anatomical scenarios such as capillary or Circle of Willis (CoW).

\paragraph{\textbf{Diffusion-based Graph Generation:}}

Recent advancements in diffusion models have led to the development of graph generation in various domains. Huang et al. \cite{huang2022graphgdp} proposed a method for generating permutation-invariant graphs. Jo et al. \cite{jo2022score} applied a score-based model for the graph generation process. Luo et al. \cite{luo2022fast} proposed a spectral denoising technique to increase the generation speed of graph models. Vignac et al. \cite{vignac2022digress} developed a method for discrete denoising diffusion for categorical node and edge variables. Concurrently, Haefeli et al. \cite{haefeli2022diffusion} also utilize discrete diffusion models for graph generation. Peng et al. \cite{peng2023moldiff} tackled the issue of atom-bond inconsistencies in 3D molecule generation. Hua et al. \cite{hua2023mudiff} proposed a unified diffusion approach for molecule generation. Vignac et al. \cite{vignac2023midi} merged discrete and continuous 3D denoising techniques to enhance molecule structure generation. Pinheiro et al. \cite{o20243d} proposed generating molecular structures within 3D spaces using voxel grids. Despite many works on molecular graph generation, there is a lack of diffusion-based vessel graph generation algorithms.

%% file: 03_method.tex
\section{Methodology}
\label{sec:meth}

We want to generate spatial graphs with nodes embedded in 3D space and edges with categorical attributes. Anatomically, the nodes represent the bifurcation points of the vascular network, and the edges represent the vessels connecting them. We denote $\boldsymbol{x}_i\in \mathbb{R}^3$ for the 3D coordinates of the $i^{\mbox{th}}$ node and $\boldsymbol{e}_{ij}\in \mathbb{R}^c$ for the edge between the $i^{\mbox{th}}$ and $j^{\mbox{th}}$ node, where $c$ is the number of edge classes. Thus, a graph is encoded by its node coordinates $\boldsymbol{X}\in\mathbb{R}^{n\times 3}$ and categorical edge adjacency matrix $\mathbf{E}\in\mathbb{R}^{n \times n \times c}$ where $n$ is the number of nodes.

\begin{figure}[t!]
    \centering
    \includegraphics[width=\linewidth]{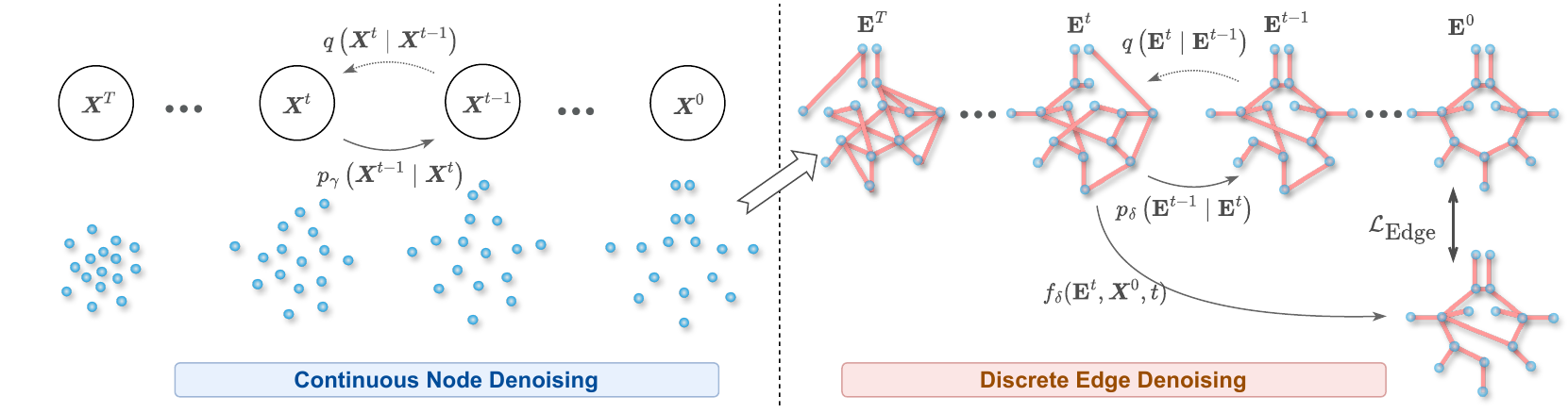}
    \caption{Overview of our method. \textbf{Left.} First, we use a continuous diffusion model to generate plausible node locations. \textbf{Right.} Subsequently, we apply discrete diffusion to generate a plausible edge configuration given the node coordinates. The node coordinates remain unchanged during edge learning. The model is optimized by $\mathcal{L}_{\mbox{Edge}}$ and focuses on the conditional edge distribution given node locations.}
    \label{fig:approach}

\end{figure}

\subsection{Two-stage Denoising}
\label{ssec:pipeline}

In typical graph domains, e.g., for molecular graphs, the atom type and rotation equivariance provide a strong bias to determine the edge class. However, a node in a vascular graph does not exhibit the same property. The vascular nodes are often assigned edge-derived attributes, such as the node degree and vessel radius, and do not necessarily contain discriminative features for edge prediction. Thus, the node location is the principal predictor of the edge existence \cite{wittmann2024link}. As a result, we believe that changing the node coordinates makes edge-type prediction difficult. Once all the node locations are known, search for plausible edge configuration benefits from the collective inductive bias of the node coordinates. In light of the above, we propose a two-stage graph generation strategy (c.f. Fig. \ref{fig:approach}). First, we focus on generating a plausible set of node coordinates as point clouds. In the second stage, we learn the edges, keeping the node coordinates fixed.
% Next, we discuss the model components in detail.

\paragraph{\textbf{Node Denoising:}}
We employ a denoising diffusion probabilistic model \cite{ho2020denoising} for the node generation task. We add Gaussian noise to the node coordinates $\boldsymbol{X}^0$ using a fixed noise model $q\left( \boldsymbol{X}^{t} \mid \boldsymbol{X}^{t-1}\right)$ as follows
\begin{equation}
    \boldsymbol{X}^{t} := \sqrt{\alpha^t} \boldsymbol{X}^{t-1} + \sqrt{1 - \alpha^t} \boldsymbol{\epsilon}; \quad \boldsymbol{\epsilon} \sim \mathcal{N}(\boldsymbol{0}, \boldsymbol{I}) \mbox{ and }  t\in [1,T]
    \label{eq:fwd_cont}
\end{equation}
\noindent where $\alpha^1, \alpha^2, \cdots, \alpha^T$ controls the noise added to the coordinates. The scheduler is selected such that $\alpha^1 \approx1$ and $\alpha^T \approx0$. This ensures that the coordinates have been mapped into a standard Gaussian.

Next, we denoise the node coordinates using a reverse diffusion process. During training, we select a random time point $t$ and train the model to predict the added noise. We employ a lightweight neural net with few fully connected layers followed by multi-head cross-attention with time embedding. We denote the model $f_\gamma$ with model parameter $\gamma$.  We obtain $\boldsymbol{X}^{t}:= \sqrt{\bar{\alpha}^t} \boldsymbol{X}^{0} + \sqrt{1 - \bar{\alpha}^t} \boldsymbol{\epsilon}$ and predict the added noise as $\boldsymbol{\hat{\epsilon}}:=f_\gamma(\boldsymbol{X}^t,t)$, where $\bar{\alpha}^t:=\prod_{s=1}^t \alpha^s$. We train $f_\gamma$ by minimizing the following loss function.
\begin{equation}
 \mathcal{L}_{\mbox{Node}} := \mathbb{E}_{t, \boldsymbol{X}^0, \boldsymbol{\epsilon}} \left\|\boldsymbol{\epsilon}-\boldsymbol{\hat{\epsilon}}\right\|^2
\end{equation}
\paragraph{\textbf{Edge Denoising:}}
Once the node denoising model is trained, we train an edge denoising model. For this, we adopt the discrete diffusion model \cite{vignac2022digress,vignac2023midi}. We start with edge attributes $\mathbf{E}^0$ and employ the following noise model. 
\begin{equation}
q\left(\mathbf{E}^t \mid \mathbf{E}^{t-1}\right):=\mathbf{E}^{t-1} \boldsymbol{Q}^t;\quad  t\in [1,T]
\label{eq:fwd_discrete}
\end{equation}
\noindent where $\boldsymbol{Q}^t_{i j}:=q\left(e^t=j \mid e^{t-1}=i\right)$ is the Markov transition probability in the state-space of edge categories. For a marginal distribution of $\boldsymbol{m}$, the resultant transition matrix is constructed as $\boldsymbol{Q}^t:=\alpha^t \boldsymbol{I}+(1-\alpha^t) \mathbf{1}_c \boldsymbol{m}^{\prime}$, where $^{\prime}$ indicates transpose operation and $\alpha^t$ is the noise scheduler similar to the node diffusion.

Following Vignac et al. \cite{vignac2023midi}, we adopt the graph transformer network \cite{dwivedi2020generalization} $f_\delta$ with parameter $\delta$ to model our edge denoising network. Note that \cite{vignac2023midi} is explicitly designed for molecular graphs and, uses rotation equivariant layers. Vascular graphs, on the other hand, can have a particular orientation based on the anatomy of interest (e.g., in the case of the Circle of Willis). A rotation equivariant graph generation model can not capture such dataset property. Hence, our model is \emph{not} rotation equivariant to preserve dataset orientation property. During training, we select a random time point $t$ and obtain $\mathbf{E}^{t}$ using Eq. \ref{eq:fwd_discrete} and predict the ground truth edge as $\mathbf{\hat{E}}^0:= f_\delta(\mathbf{E}^t,\boldsymbol{X}^0, t)$. The model is supervised by cross-entropy (CE) loss.
\begin{equation}
  \mathcal{L}_{\mbox{CE}} := \mathbb{E}_{t, \mathbf{E}^0} \mathbb{E}_{q\left(\mathbf{E}^t \mid \mathbf{E}^0\right)} \sum_{ij} CE(\mathbf{E}^0_{ij},\mathbf{\hat{E}}^0_{ij})
\end{equation}
In addition to the correct edge class, the node degree between edges is vital in vascular graphs \cite{lyu2022reta}. Thus, we introduce a novel node degree loss. With multiple valid edge configurations for the given node locations, the model should predict a degree distribution akin to the ground truth. Thus, we use a KL divergence loss between prediction and target node degree distribution over a mini-batch. However, we need the adjacency matrix to compare the node degree, which requires discrete sampling from the predicted edge adjacency likelihood. This operation would break the gradient backpropagation. To address this issue, we use the Gumbel-softmax ($GS$) trick for the adjacency matrix sampling. The node degree loss is computed as
\begin{equation}
 \mathcal{L}_{\circ} := KL\left(p_{\mbox{deg}}(\mathbf{E}^0) || p_{\mbox{deg}}(GS(\mathbf{\hat{E}}^0))\right)
\end{equation}
The total loss for edge denoising is $\mathcal{L}_{\mbox{Edge}}=\mathcal{L}_{\mbox{CE}}+\mathcal{L}_\circ$. We study the contribution of these losses in our ablation study.

\subsection{Graph Generation}
Once both $f_\gamma$ and $f_\delta$ are trained, we can use it for sampling vessel graphs. First, we sample the number of nodes, $n$, from possible discrete values. Next, we sample $\boldsymbol{X}^T\sim \mathcal{N}(\boldsymbol{0}, \boldsymbol{I})$ and perform the following denoising steps from $t=T,T-1,\cdots,1$ to obtain the posterior $p_\gamma\left(\boldsymbol{X}^{t-1}\mid \boldsymbol{X}^{t}\right)$ as follows:
\begin{equation}
    \boldsymbol{X}^{t-1} = \frac{1}{\sqrt{\alpha^t}}\left(\boldsymbol{X}^t-\frac{1-\alpha^t}{\sqrt{1-\bar{\alpha}^t}} \boldsymbol{\hat{\epsilon}}^t\right) + \sqrt{1 - \alpha^{t}} \boldsymbol{\epsilon}
\end{equation}

\noindent where $\boldsymbol{\epsilon} \sim \mathcal{N}(\boldsymbol{0}, \boldsymbol{I})$ if $t>1$, else $\boldsymbol{\epsilon}=\boldsymbol{0}$. Once we have $\boldsymbol{X}^0$, we sample $\mathbf{E}^T_{ij}\sim \boldsymbol{m}$ to start the edge denoising steps from $t=T,T-1,\cdots,1$. We first obtain $\mathbf{\hat{E}}^0$ from $\mathbf{E}^t$ and compute the posterior distribution of the edge category using the following as proposed in \cite{austin2021structured,vignac2022digress}
\begin{equation}
    p_\delta\left(\mathbf{E}^{t-1}_{ij}\mid\mathbf{E}^t_{ij}\right)= \frac{\mathbf{E}^t_{ij} {\boldsymbol{Q}^t}^{\prime}_{ij} \odot \mathbf{\hat{E}}^0_{ij} \bar{\boldsymbol{Q}}^{t-1}_{ij}}{\mathbf{\hat{E}}^0_{ij} \bar{\boldsymbol{Q}}^t_{ij} {\mathbf{E}^t}^{\prime}_{ij}}
\end{equation}

\noindent where $\bar{\boldsymbol{Q}}^t:=\boldsymbol{Q}^1 \ldots \boldsymbol{Q}^t$ and $\odot$ denotes pointwise multiplication. In the next step, we sample the categorical value of edge attributes from $\mathbf{E}^{t-1}_{ij} \sim p_\delta\left(\mathbf{E}^{t-1}_{ij}\mid\mathbf{E}^t_{ij}\right)$.

%% file: 04_exp.tex
\section{Experiments}
\label{sec:exp}

\paragraph{\textbf{Datasets:}}
% \label{ssec:data}
\begin{figure}[t!]
    \centering
    \includegraphics[width=\textwidth]{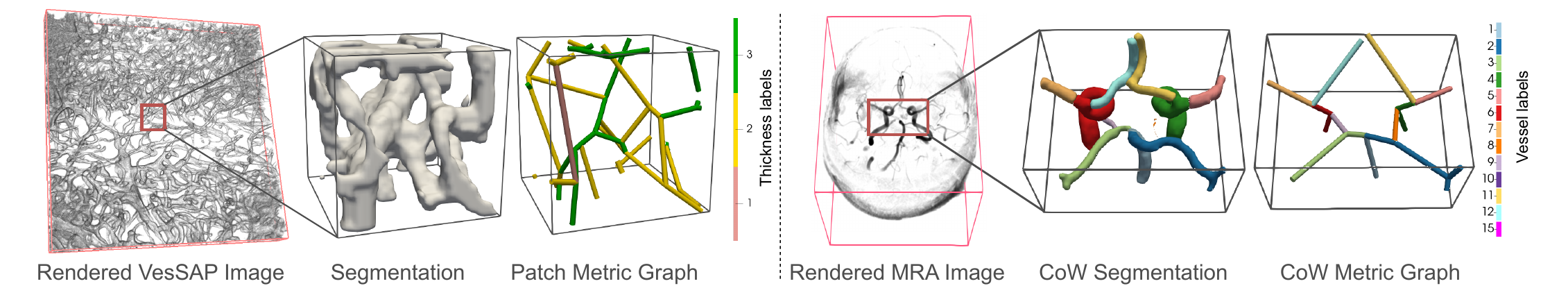}
    \caption{Two real-world vessel datasets. \textbf{Left.} An example from VesSAP and the capillary level vessel graph patch. \textbf{Right.}The extracted metric graph for the blood vessel in CoW. (Best viewed when zoomed in) }
    \label{fig:dataset}
\end{figure}

We have experimented on two real-world publicly available datasets. The first one consists of microscopic images of capillary-level vessel graphs, namely VesSAP \cite{todorov2020machine,paetzold2021whole}. We have chosen 24 annotated volumes, and the extracted graphs from Voreen \cite{drees2021scalable}. For our experiments, we used a patch size graph of $48\times 48\times 48$ in the voxel space in the original resolution. We have used the radius information to create 4 edge classes based on thickness, including one background (no-edge) class. For the second dataset, we have the Circle of Willis (CoW) dataset consisting of 390 MRA images from CROWN \footnote{\url{https://crown.isi.uu.nl/}} (n=300) and TopCoW (n=90) \cite{yang2023benchmarking} challenges. We used a multi-class segmentation tool trained on the TopCoW dataset to segment the CROWN datasets and Voreen to extract graphs from the segmentation. For CoW data, we used the whole CoW graph as one sample, and it has 14 classes of edges containing different artery labels and a background class. For both datasets, we have used the refined graph \cite{drees2021scalable}.

\paragraph{\textbf{Baselines:}} There is a notable void in data-driven vessel graph generation methods. Although tree-generation methods exist, they are not applicable in our setting since we are also interested in generating cycles in the graph. Therefore, we select two recent molecular graph generation methods, namely Congress \cite{vignac2022digress} and MiDi \cite{vignac2023midi}, that can generate 3D spatial graphs with cycles as baselines.

\paragraph{\textbf{Metrics:}}
% \label{ssec:metrics}

We acknowledge that finding a single good metric to benchmark vessel graph generation methods combining both the node coordinates and the edge information is difficult. It has been reported that when the generated graphs accurately represent real data distribution, they improve downstream performance \cite{kreitner2024synthetic}. Previous work \cite{feldman2023vesselvae} has used distributional differences in radius, total length, and tortuosity between generated and real mesh. Along this line, we adopt the following physical parameters to compute the KL divergence between real and generated graphs. For nodes, we use 3D coordinate positions (\textit{$\{x,y,z\}$}) and node degree (\textit{deg($\mathcal{V}$)}). For edges, we use total edge number (|$\mathcal{E}$|), edge length ($l_\mathcal{E}$), angle between edges ($\mathcal{E} ~\angle$), edge orientation with three axes \textit{$\{\theta,\phi,\psi\}$}, Bett-0 for connected component (\textit{$\beta_0$}), and Betti-1 for cycles (\textit{$\beta_1$}).

\paragraph{\textbf{Implementation Details:}}
% \label{ssec:implmentation} 

The node-denoising network works with normalized coordinates and uses two blocks, each containing 1) A multi-layer perception (MLP) to process the coordinates, 2) an MLP to process the denoising timestep (encoded as a sinusoidal positional embedding), and 3) A multi-head self-attention (MSA) processes the summed output of the two MLPs. A final MLP projects the features to $\mathbb{R}^3$. All the layers use a hidden dimension of size 256.
The edge-denoising network is composed of eight blocks. Each block gets the node coordinates, noisy edges, and the timestep information as input. A feature transformation layer (FiLM)~\cite{perez2018film} processes the edges and timestep information and merges them with node features. The resultant node features are passed through an MSA. The final node features predict the output edges. Please refer to the supplementary for all models' hyperparameters.
All models (including baselines) are trained from scratch for 1000 epochs on a single A-6000 GPU with a batch size 64. We use AdamW~\cite{loshchilov2017decoupled} optimizer with a learning rate of 0.0003. We implement the code using the pyTorch and pyTorchGeometric libraries. Our code is available at \url{https://github.com/chinmay5/vessel_diffuse}

\begin{table*}[t!]%[htpb]

  \caption[table: Comparison]{ Comparison of our method against the existing spatial graph generation methods. Our proposed two-stage solution outperforms the existing baselines, achieving the lowest KL divergence on both the VesSAP and the CoW datasets.}
  \label{tab:results_SOTA}
  \centering
  \resizebox{\textwidth}{!}{
  \setlength{\tabcolsep}{2mm}{
   \begin{tabular}{l |l | c | c | c| c | c| c | c | c}
    \toprule
     & {Methods} &\textit{$x,y,z$} &\textit{deg($\mathcal{V}$)}  & |$\mathcal{E}$| &  $l_\mathcal{E}$ & $\mathcal{E} ~\angle$& \textit{$\theta,\phi,\psi$}&\textit{$\beta_0$} &\textit{$\beta_1$}\\
    \midrule
     \multirow{3}*{\rotatebox[origin=c]{90}{VesSAP}} & Congress \cite{vignac2022digress} & 0.003 & 0.211 & 0.288 &0.112 & 0.202 & 0.002 & 6.226 & 8.251\\
     & MiDi \cite{vignac2023midi} & 0.011 &0.023 &0.002 & \textbf{0.003} & 0.011 & 0.003 & 0.014 & 0.074\\
    & \textbf{Ours} &\textbf{0.002} &\textbf{0.006} & \textbf{0.001} & 0.006 & \textbf{0.005} & \textbf{0.001} & \textbf{0.001} & \textbf{0.068} \\
    \midrule
    % CoW
     \multirow{3}*{\rotatebox[origin=c]{90}{CoW}} & Congress \cite{vignac2022digress} & 0.003 & 0.010 & 0.046 & 0.065 & 0.081 & 0.005 & 0.051 & 0.032 \\
     & MiDi \cite{vignac2023midi} & 0.010 & 0.011 & 0.040 & 0.012 & \textbf{0.007} & 0.005 & 0.049 & 0.028\\
    & \textbf{Ours} & \textbf{0.002} &  \textbf{0.003} & \textbf{0.004} & \textbf{0.006} & \textbf{0.007} & \textbf{0.002} &\textbf{0.015} & \textbf{0.017}\\

    \bottomrule
  \end{tabular}
}
}
\end{table*}

% \subsection{Results and Discussion}
\paragraph{\textbf{Results and Discussion:}}
\label{ssec:results}

Table ~\ref{tab:results_SOTA} compares the statistics of the generated 3D vessel graphs with the state-of-art methods, viz., the discrete graph generation model MiDi and the continuous model, Congress. Congress yields overconnected graphs, failing to reflect the actual distribution. MiDi performs better but struggles with connectivity (node degrees, $\beta_0$) and edge properties (length, number of edges). Our graph generation model outperforms the baseline methods on both CoW and VesSAP datasets, producing graphs that most closely emulate the sample statistics and obtain the lowest KL divergence for various graph and topology metrics. This result validates the utility of our two-stage design principle, which starts with node diffusion and then moves on to edge diffusion. Please refer to the supplementary for a detailed comparison of distribution.

\begin{figure}[t!]
  \centering
  \includegraphics[width=\textwidth]{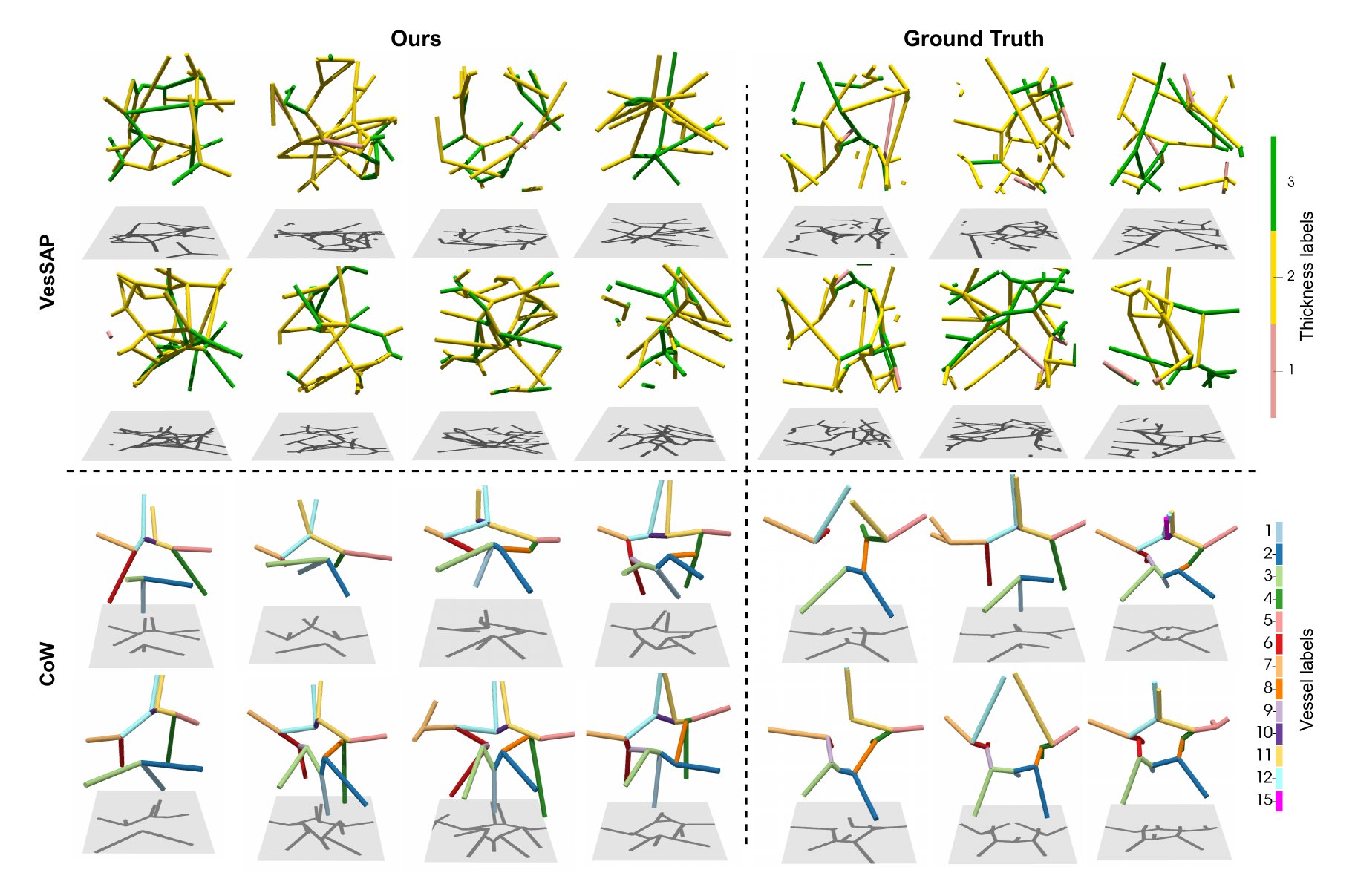} 

  \caption{ VesSAP and CoW graphs generated by our model in comparison to ground truth samples. Our model is able to learn complex structures, such as loops and orientation characteristics for both datasets.}
  \label{fig:placeholder}

\end{figure}

Figure \ref{fig:placeholder} shows visual examples of the graphs generated by our model and the ground truth. Our model is able to capture the complex vascular shapes in the VesSAP dataset and can generate cycles of similar complexity. The CoW dataset has a specific orientation of edges, which is again captured faithfully by our model. Please refer to the supplementary for more qualitative results.

\begin{table*}[!bt]%[htpb]
  \caption[table: Comparison]{ Ablation of rotation equivariance (R.E.), two-stage denoising model (II S), and the degree loss $\mathcal{L}_{\circ}$ for CoW dataset. The best model is not rotation equivariant and uses two-stage denoising and degree loss. All values are reported in units of $10^{-2}$.}
  \label{tab:ablation}
  \centering
  \resizebox{\textwidth}{!}{
  \setlength{\tabcolsep}{3mm}{
   \begin{tabular}{l | l |l | c | c| c| c | c| c | c | c}
    \toprule
 
    R.E. & II S & $\mathcal{L}_{\circ}$ &\textit{$x,y,z$} &\textit{deg($\mathcal{V}$)}  & |$\mathcal{E}$| & $l_\mathcal{E}$ & $\mathcal{E} ~\angle$& \textit{$\theta,\phi,\psi$}&\textit{$\beta_0$} &\textit{$\beta_1$}\\

    \midrule

    \cmark & \xmark & \xmark & 1.04 & 1.12 & 4.01 & 1.29 & 0.78 & 0.52 & 4.89 & 2.01\\
     
     \xmark & \xmark & \xmark & 0.44 & 1.89 & 4.43 & 6.08 & 8.14 & 0.33 & 11.16 & 4.31\\ 
     
     \xmark  & \cmark  &\xmark & \textbf{0.21} & \textbf{0.29} & 3.43 & 0.74 & 0.81 & 0.19 & \textbf{1.10} & 2.31 \\
     
     \xmark  & \cmark  &\cmark & \textbf{0.21} &  \textbf{0.29} & \textbf{0.40} & \textbf{0.63} & \textbf{0.74} & \textbf{0.19} &1.54 & \textbf{1.71}\\
    
    \bottomrule
  \end{tabular}
}
}
\end{table*}

\paragraph{\textbf{Ablation:}} We analyze the importance of different components of our proposed method. We choose the Circle of Willis dataset for our ablation experiments. We summarize the results in Table ~\ref{tab:ablation}.
Disabling the rotation equivariance (R.E.) in the edge denoising blocks enables the model (Row 2) to learn about the vascular graphs' inherent orientation bias. Although this leads to graphs with better orientation ($\theta, \phi, \psi$), R.E. is a powerful inductive bias for edge connectivity, and naively turning it off leads to very sparse graphs (significant deterioration in $\beta_0$).
We switch to our two-stage (II S) denoising approach to counteract the increase in sparsity. 
We observe an immediate performance improvement, with most statistics better than the baseline (Row 3 \emph{vs.} Row 1).
We further improve the quality of the generated samples by incorporating a loss on the node degree distribution. This loss provides additional supervision for generating graphs and further improves topological properties, leading to samples that most closely emulate the ground truth distributions (Row 4).

%% file: 05_supp.tex
\section*{}
\begin{figure}[!ht]
\centering
		\includegraphics[trim=50 40 40 70 , clip, width=0.99\textwidth]{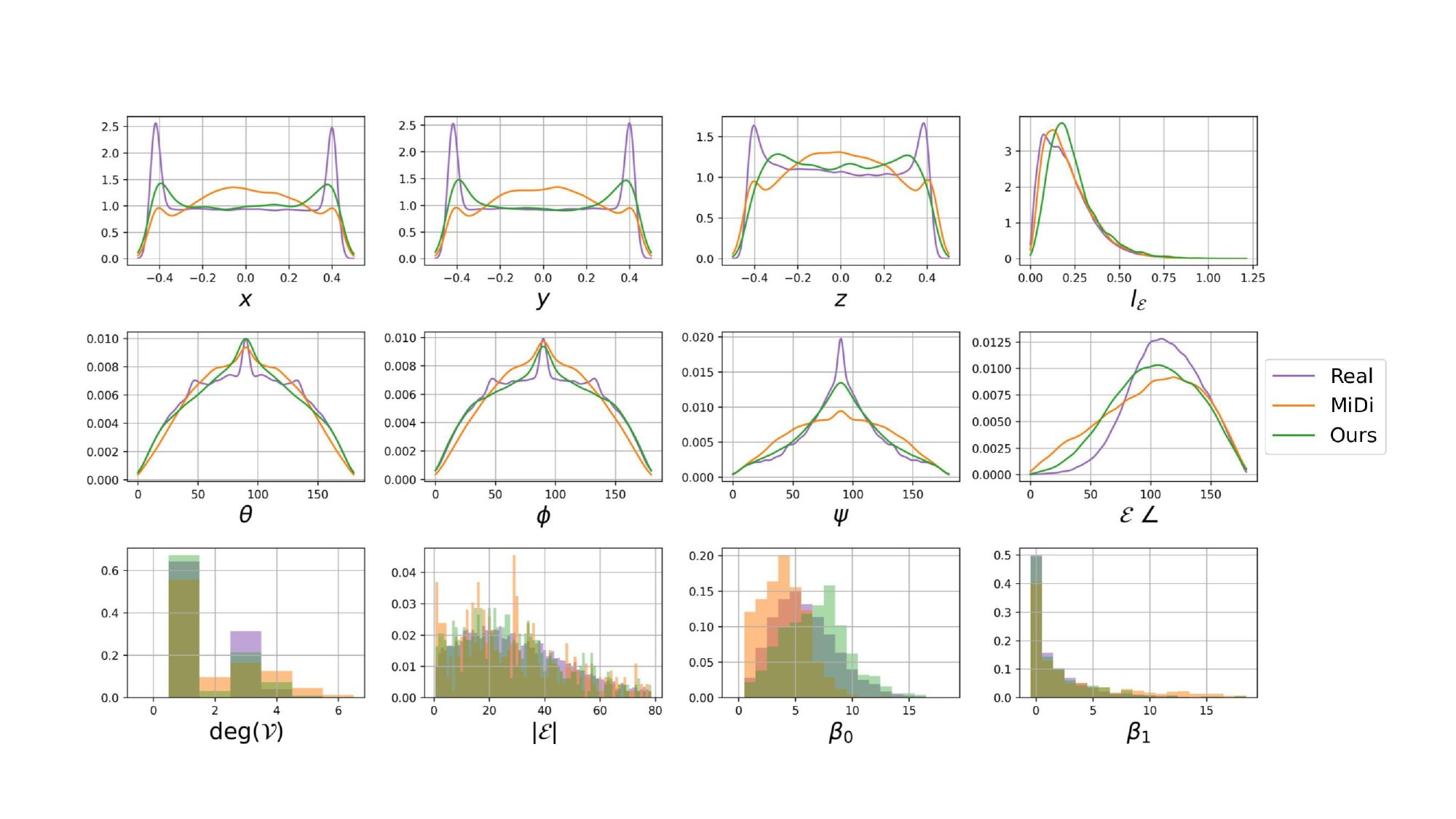}

 (a) VesSAP Dataset
		\includegraphics[trim=50 40 40 70 , clip, width=0.99\textwidth]{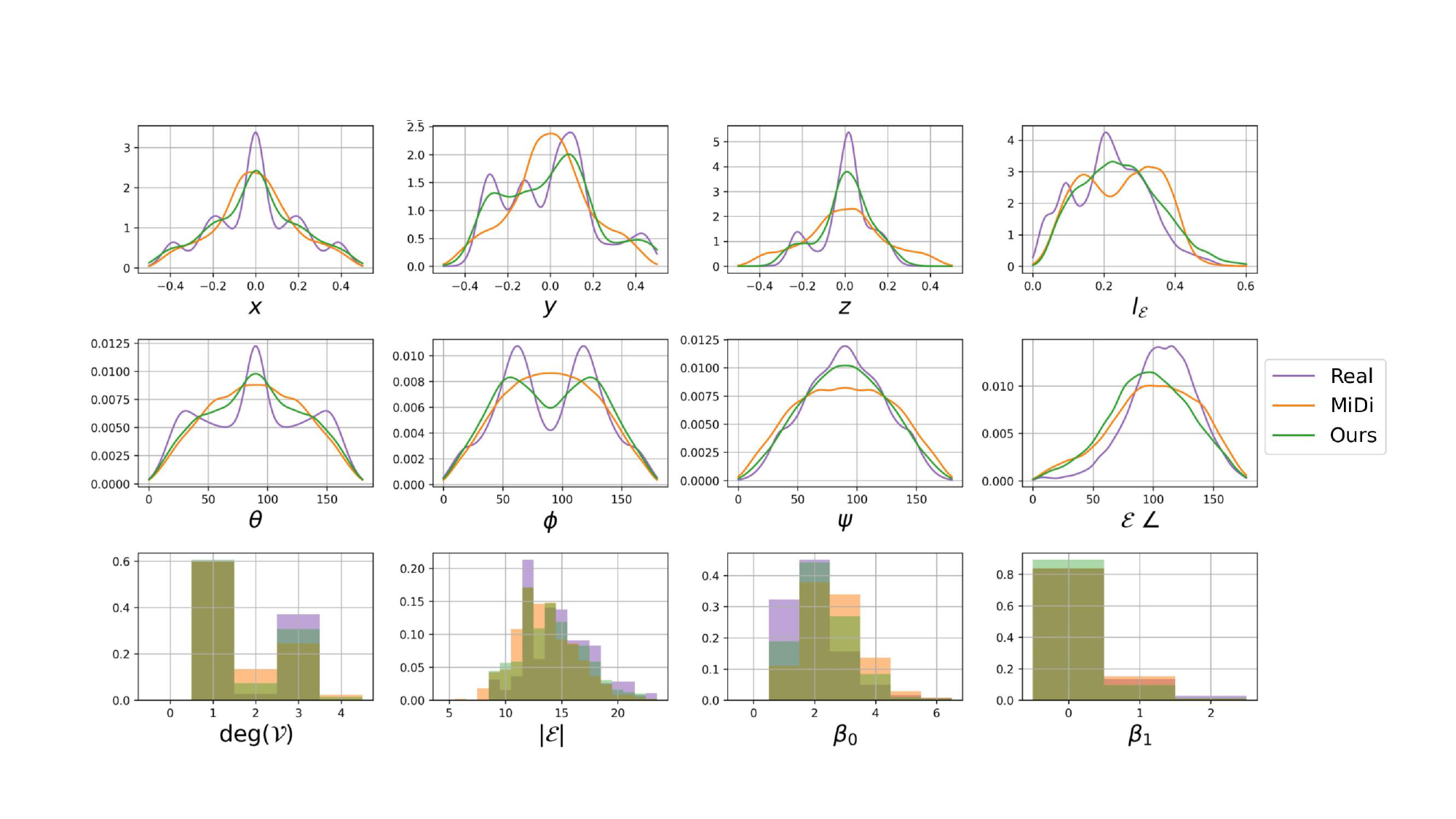}
   (b) CoW Dataset
    \caption{Comparison of our method against the MiDi for the VesSAP and Circle of Willis (CoW) datasets. We show the results for learning coordinates (\textit{$x,y,z$}), edge angles ($\mathcal{E} ~\angle$), edge orientation (\textit{$\theta,\phi,\psi$}) and edge length ($l_\mathcal{E}$). We also compare the statistics for node degree(\textit{deg($\mathcal{V}$)}), number of edges (|$\mathcal{E}$|), and the Betti values (\textit{$\beta_0$} \textit{$\beta_1$}). The ground truth distribution is in purple, the distribution learned by MiDi is in orange, and our method is in green. As observed, our method emulates the statistics of the ground truth graphs more faithfully than MiDi. The degree of distribution of MiDi vs. our method on the VesSAP dataset is especially interesting. While the VesSAP graphs contain no degree 2 nodes, MiDi generates graphs with a large number of degree 2 nodes. Our method overcomes this shortcoming and generates minimal degree 2 nodes.}
	\label{fig:dist_crown} 
\end{figure}

\begin{figure}[t!]
  \centering
  \includegraphics[width=\textwidth]{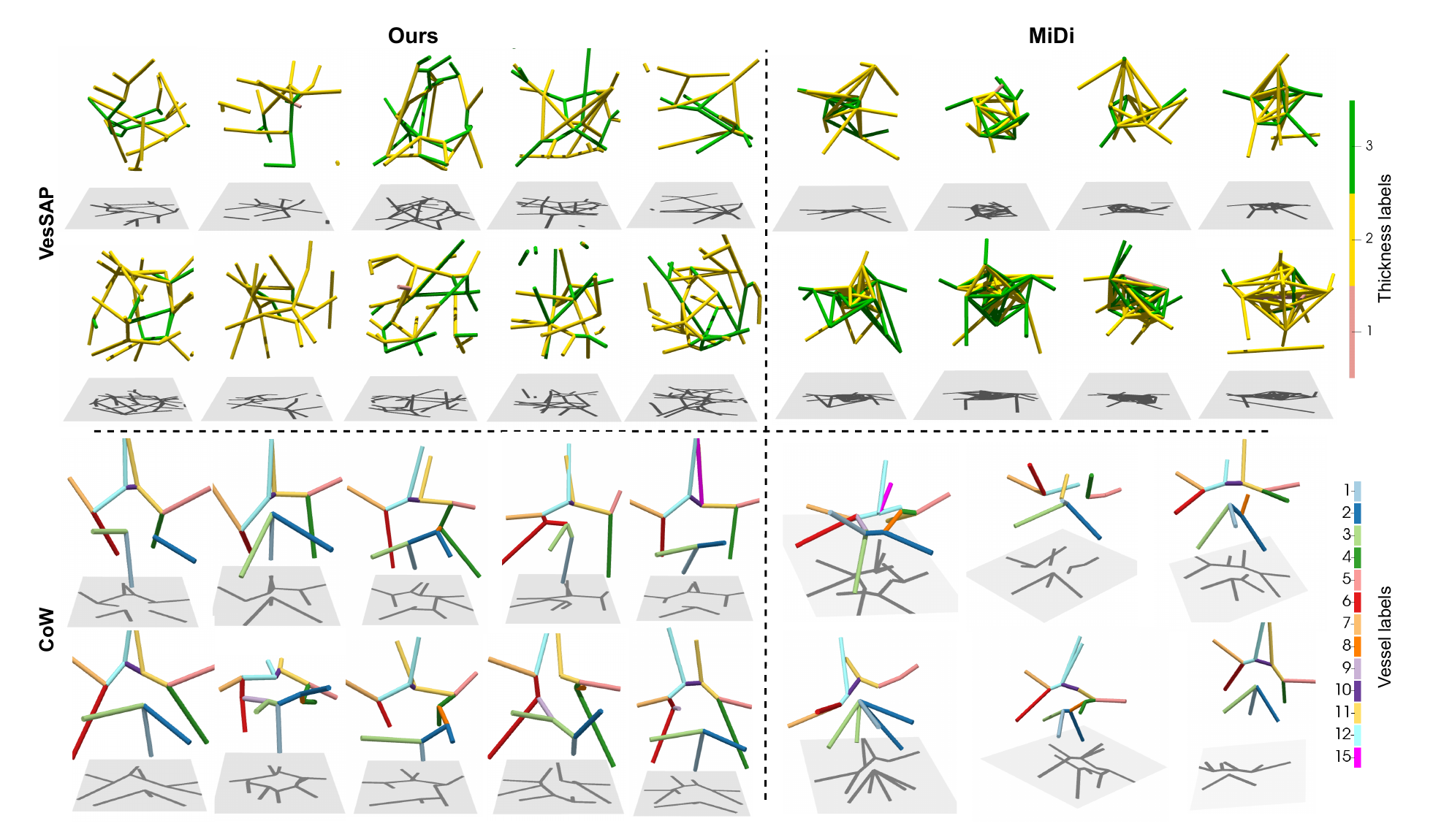} 
  \caption{VesSAP and CoW graphs generated by our model and MiDi, respectively. Note that, in the case of VesSAP, MiDi failed to capture the node coordinate distribution, which is the driving property for correct edge distribution, and hence produces an overconnected graph, resulting in a high Betti number error. However, MiDi fares relatively well in the CoW configuration, which resembles a molecular layout. In contrast, our model is able to generate diverse and valid novel graphs for both datasets.}
  \label{fig:placeholder1}
\end{figure}

\begin{table*}[th]%[htpb]
\centering
\caption[table: hyper_params]{ Details of the hyper-parameters used by the baselines (Congress and MiDi) and our method. We report the number of transformer blocks used by the models, the number of attention heads in each transformer block, the number of denoising steps, and the node and edge projection dimensions. We report the parameters of our edge denoising and coordinate diffusion modules in separate columns for clarity. We have tried to use a similar number of trainable parameters for all three methods for a fair comparison.}
\label{tab:hyper}
\resizebox{0.7\textwidth}{!}{
\setlength{\tabcolsep}{2mm}{
\begin{tabular}{l|c|c|c}
\midrule
 \multirow{2}{*}{Model Specification} &  Congress\cite{vignac2022digress} &  \multicolumn{2}{c} {Ours}\\
 \cline{3-4}
& MiDi\cite{vignac2023midi}    &  Edge Network & Node Network\\
\toprule
Model Type & Transformer & Transformer & MLPs+MSA \\
\# blocks                     & 8      & 8    & 2\\ 
\# attention heads                              & 8      & 8    & 4  \\ 
\#  node features                     & 128      & 128    & 256  \\ 
\#  edge features                     & 64       & 64     & - \\ 
\#  denoising steps                    & 1000     & 1000    & 1000 \\ 

 \bottomrule
\end{tabular}
}
}

\end{table*}